\documentclass[letterpaper]{article} 
\usepackage{aaai2026}  
\usepackage{times}  
\usepackage{helvet}  
\usepackage{courier}  
\usepackage[hyphens]{url}  
\usepackage{graphicx} 
\usepackage{natbib}  
\usepackage{caption} 
\usepackage{algorithm}
\usepackage{algorithmic}
\usepackage{subcaption}     
\usepackage{booktabs}
\usepackage{amsmath}
\usepackage{newfloat}
\usepackage{listings}
\DeclareCaptionStyle{ruled}{labelfont=normalfont,labelsep=colon,strut=off} 
\floatstyle{ruled}
\newfloat{listing}{tb}{lst}{}
\floatname{listing}{Listing}
\title{Agentmandering: A Game-Theoretic Framework for Fair Redistricting via Large Language Model Agents}
\author {
    Hao Li\textsuperscript{\rm 1}\equalcontrib,
    Haotian Chen\textsuperscript{\rm 2}\equalcontrib,
    Ruoyuan Gong\textsuperscript{\rm 1},
    Juanjuan Wang\textsuperscript{\rm 3},
    Hao Jiang\textsuperscript{\rm 1}\thanks{Corresponding author},
}
\affiliations{
    \textsuperscript{\rm 1}Wuhan University, Wuhan, 430072 China\\
    \textsuperscript{\rm 2}University of California, Los Angeles, Los Angeles, 90095 \\
    \textsuperscript{\rm 3}Zhongnan University of Economics and Law, Wuhan, 430073 China\\

    whulh@whu.edu.cn, barneychen@ucla.edu, GongRuoyuan@whu.edu.cn, Wangjj@zuel.edu.cn, jh@whu.edu.cn\\
}

\begin{document}

\maketitle

\begin{abstract}
    Redistricting plays a central role in shaping how votes are translated into political power. While existing computational methods primarily aim to generate large ensembles of legally valid districting plans, they often neglect the strategic dynamics involved in the selection process. This oversight creates opportunities for partisan actors to cherry-pick maps that, while technically compliant, are politically advantageous. Simply satisfying formal constraints does not ensure fairness when the selection process itself can be manipulated. We propose \textbf{Agentmandering}, a framework that reimagines redistricting as a turn-based negotiation between two agents representing opposing political interests. Drawing inspiration from game-theoretic ideas, particularly the \textit{Choose-and-Freeze} protocol, our method embeds strategic interaction into the redistricting process via large language model (LLM) agents. Agents alternate between selecting and freezing districts from a small set of candidate maps, gradually partitioning the state through constrained and interpretable choices. Evaluation on post-2020 U.S. Census data across all states shows that Agentmandering significantly reduces partisan bias and unfairness, while achieving 2 to 3 orders of magnitude lower variance than standard baselines. These results demonstrate both fairness and stability, especially in swing-state scenarios. Our code is available at \url{https://github.com/Lihaogx/AgentMandering}.
\end{abstract}

\section{Introduction}
\begin{figure*}[ht!]
    \centering
    \begin{subfigure}[b]{0.49\textwidth}
        \centering
        \includegraphics[width=\textwidth]{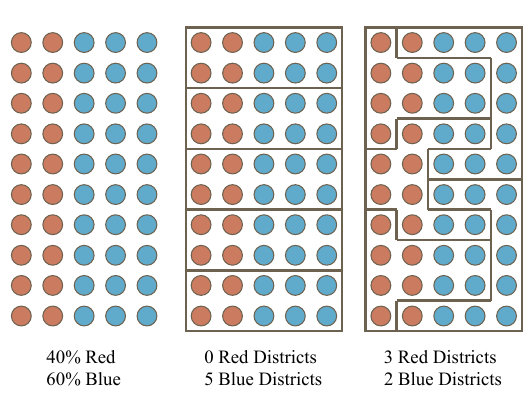}
        \caption{}
        \label{fig:gerrymandering}
    \end{subfigure}
    \hfill
    \begin{subfigure}[b]{0.49\textwidth}
        \centering
        \includegraphics[width=\textwidth]{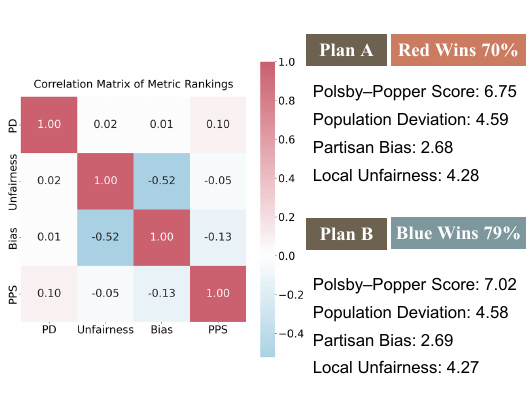}
        \caption{}
        \label{fig:unfair}
    \end{subfigure}
    \caption{(a) Gerrymandering example showing district manipulation. (b) Correlation analysis of four redistricting metrics and comparison of two districting plans.}
    \label{fig:redistricting}
\end{figure*}

In representative democracies, electoral districts determine how citizens are grouped for political representation. In the United States, the winner-takes-all and single member system makes electoral results highly sensitive to district boundaries, which significantly affect the results of congressional and state legislative races~\cite{cox2002elbridge}. Redistricting, the periodic redrawing of district boundaries to reflect population changes, is essential for equitable representation. However, this process is frequently manipulated for political advantage through a practice known as partisan \textit{gerrymandering}, where district lines are intentionally designed to favor the party that controls the drawing~\cite{Gelman1994-sg}. Common tactics include \textit{packing} voters into a small number of districts to concentrate their influence, or \textit{cracking} them across many districts to dilute their voting power. As illustrated in Figure~\ref{fig:gerrymandering}, a blue-majority population can be divided such that the red-minority secures more districts, demonstrating a classic gerrymandering situation that undermines fair representation.

Modern computational approaches of redistricting focus on generating large ensembles of districting plans that comply with formal legal and demographic constraints. Methods such as Markov Chain Monte Carlo (MCMC)~\cite{recom, Chikina2017-ek, Carter2019-ge}, Sequential Monte Carlo (SMC)~\cite{McCartan2020-uz}, and integer programming~\cite{fravel2023optimizing} are commonly used to produce thousands of plausible alternatives, enabling statistical comparisons to identify instances of extreme partisan bias. However, as shown in Figure~\ref{fig:unfair}, the heatmap reveals that the four key evaluation metrics exhibit low correlations, suggesting that these metrics capture orthogonal dimensions of fairness. This independence creates opportunities for partisan manipulation: as shown in the example, two plans with nearly identical metric scores can lead to starkly different electoral outcomes~\cite{chambers2017flaws, Barnes2021-sq}. In practice, the abundance of legally compliant maps can be exploited by political actors who select technically valid plans that subtly serve partisan goals. Thus, simply generating maps that meet formal constraints is insufficient. The central challenge is \textbf{how to generate plans that are not only compliant, but also robust against strategic selection and capable of achieving fair outcomes under adversarial decision-making}.

Recent work in fair redistricting has proposed negotiation-based protocols that aim to achieve equitable outcomes through structured interaction. One prominent example is the \textit{Choose-and-Freeze} protocol~\cite{Pegden2017-az}, which draws from classical ideas in fair division and game theory, particularly the ``cake-cutting'' paradigm~\cite{brams2006better}. In this game-theoretic setting, two opposing parties take turns: one selects a complete districting plan, and the other freezes a single district from it. The process then recurses on the remaining territory. This alternating structure creates a balanced strategic environment in which each side possesses both agency and constraint. The protocol has been proven to produce fair outcomes under reasonable assumptions, offering formal guarantees of envy-freeness and symmetry without relying on external arbiters or optimization objectives.

Despite these theoretical advantages, protocols like \textit{Choose-and-Freeze} remain largely disconnected from computational practice. They are designed to guide human negotiation, but cannot be directly implemented using existing algorithmic redistricting pipelines. Most modern methods focus on generating large ensembles of legal plans via sampling or optimization, lacking the interactive structure and strategic balance that these protocols embody.

To bridge the gap between theoretical negotiation protocols and practical redistricting methods, we introduce \textbf{Agentmandering}, a framework that implements the game-theoretic \textit{Choose-and-Freeze} protocol~\cite{Pegden2017-az} using large language model (LLM) agents~\cite{li2025political}. At each step, a small set of feasible districting plans is generated over the remaining unpartitioned region. One agent chooses a preferred map, and the opposing agent freezes a single district from it. The process then recurses until the full state is partitioned. By leveraging the LLMs' capacity for strategic reasoning and preference modeling, we simulate partisan decision-making within a structured, bounded-interaction protocol. This design enables the practical realization of theoretically fair procedures via AI agents, yielding districting outcomes that are both interpretable and robust to strategic manipulation.

\begin{enumerate}
    \item We introduce a new paradigm that leverages LLM agents to implement game-theoretic protocols, bringing abstract fairness principles into practical tools for computational redistricting.
    \item We introduce \textbf{Agentmandering}, a framework combining the \textit{Choose-and-Freeze} protocol with LLM agents to structure redistricting as a strategic negotiation, constraining partisan manipulation and yielding fairer outcomes.
    \item We demonstrate that Agentmandering achieves 2 to 3 orders of magnitude lower variance than existing methods on post-2020 U.S. Census data across all states, while reducing partisan bias and unfairness. This highlights its robustness across all states, particularly in swing-state scenarios.
\end{enumerate}

\subsection{Related Work}

\paragraph{Evaluation and selection of districting plans.} Previous research by mathematicians, computer scientists, and legal scholars has pursued two main approaches to combat gerrymandering: 1) developing a metric (such as \textit{efficiency gap} and \textit{compactness}) to evaluate and optimize fairness across large collections of redistricting plans from simulations~\cite{Niemi1990-mw, stephanopoulos2015partisan, Ko2022-lt}; and 2) designing map-drawing algorithms to ensure overall partisan fairness. The first approach, however, depends on judicial rulings in partisan gerrymandering cases---after the federal courts' withdrawal in \textit{Rucho v. Common Cause}, this path remains deeply contested~\cite{Chen2015-yi, Tam_Cho2016-jl}. Our research aligns with the second approach and contributes to a growing body of work in which scholars propose interactive protocols that partition maps through negotiations between opposing parties~\cite{landau2009fair, Pegden2017-az, mixon2018utility, benade2023you, palmer2024partisan}, analogous to the classic ``cake-cutting'' problem~\cite{brams2006better}. This approach does not rely on independent commissions or special masters to draw the map. Instead, each party acts in its own interest and takes turns making mapping decisions until they reach an unique subgame perfect equilibrium. 

\paragraph{Game-theoretic LLM agents in negotiations.}
With the emergence of LLM agents, a growing body of research has explored whether LLMs can engage in strategic behavior through autonomous reasoning in negotiation settings. Several studies use classic game-theoretic environments to create controlled settings for evaluating LLM agents' human-like strategic interaction~\cite{guo2023GPT, mao2024alympics, fan2024can, gemp2024steering, hua2024game}. Other work applies LLMs to real-world social deduction games such as Avalon~\cite{light2025strategist}, Werewolf~\cite{xu2024exploring}, and Chameleon~\cite{karabag2025llm}, examining whether agents can navigate complex rule-based behavior similar to humans. A parallel line of research investigates LLM agents in realistic economic contexts, such as trade and auction decision-making~\cite{jiang2025harbor, kwon2025astra}, to test their ability to reason under market constraints. Building on this literature, our work extends the use of LLM-based agents to the high-stakes domain of political fairness, applying strategic negotiation to the problem of electoral redistricting.

\section{Preliminaries}

\paragraph{Redistricting}

Redistricting is the process of redrawing electoral district boundaries to reflect population changes and maintain fair political representation, typically carried out after each decennial census. In computational terms, it is often formulated as a graph partitioning problem. Let \( G = (V, E) \) denote the adjacency graph of population units (e.g., precincts or census blocks), where each node \( v \in V \) is assigned a population weight \( p(v) \). The task is to partition \( V \) into \( k \) disjoint subsets \( \{V_1, V_2, \dots, V_N\} \), each representing a district, subject to the following conditions. Each district \( V_i \) must induce a connected subgraph of \( G \) (contiguity), and the population must be balanced such that
\[
\left| \sum_{v \in V_i} p(v) - \frac{1}{N} \sum_{v \in V} p(v) \right| \leq \epsilon,
\]
for a given tolerance \( \epsilon \). Additional constraints may also apply, including geometric compactness, preservation of communities of interest, or compliance with legal mandates such as the Voting Rights Act.

\paragraph{Metrics of Redistricting Evaluation}
\label{sec:metrics}

Given a districting plan \( \mathcal{M} = \{V_1, V_2, \dots, V_N\} \) over a population graph \( G = (V, E) \), we evaluate its quality using the following key metrics:

\textit{Population Deviation (PD)} measures how equally population is distributed across districts~\cite{stephanopoulos2015partisan}. It is defined as the average deviation from the ideal district population:
\[
\mathrm{PD} = \frac{1}{N} \sum_{i=1}^{N} \left| \sum_{v \in V_i} p(v) - \frac{1}{N} \sum_{v \in V} p(v) \right|.
\]

\textit{Polsby--Popper Score (PPS)} quantifies the geometric compactness of a district~\cite{polsby1991third}. For each district \( V_i \), it is computed as:
\[
\mathrm{PPS}(V_i) = \frac{4\pi A_i}{P_i^2},
\]
where \( A_i \) and \( P_i \) are the area and perimeter of district \( V_i \), respectively. We report both the average and minimum PPS over all districts.

\textit{Partisan Bias (Bias)} captures systemic advantage for one party over another~\cite{grofman2007future}. It is calculated as the average deviation from parity between vote share and seat share:
\[
\mathrm{Bias} = \frac{1}{N} \sum_{i=1}^{N} (2 \cdot \mathrm{pct\_dem}_i - 1),
\]
where \( \mathrm{pct\_dem}_i \) is the Democratic vote share in district \( V_i \).  
Values closer to 0 indicate fairer partisan balance. A positive value suggests a bias in favor of the Democratic Party (i.e., districts are drawn to favor Democrats), while a negative value indicates a bias in favor of the Republican Party.

\textit{Unfairness} measures the proportion of residents whose preferred party did not win in their district~\cite{ko2022all}. For each district \( V_i \) with population \( P_i \), we define:
\[
\mathrm{unhappy\_votes}_i =
\begin{cases}
(1 - \mathrm{pct\_dem}_i) \cdot P_i & \text{if Dem win}, \\
\mathrm{pct\_dem}_i \cdot P_i & \text{otherwise}.
\end{cases}
\]
And the overall unfairness is computed as:
\[
\mathrm{Unfairness} = \frac{\sum_i \mathrm{unhappy\_votes}_i}{\sum_i P_i}.
\]

Lower \textit{PD} and \textit{Unfairness} indicate better population balance and greater voter satisfaction, higher \textit{PPS} implies more compact districts, and smaller absolute \textit{Bias} signals reduced partisan skew.

\paragraph{Gerrymandering}
Gerrymandering refers to the manipulation of electoral district boundaries to favor a specific party or group. The term dates back to 1812, when a Massachusetts district approved by Governor Elbridge Gerry was said to resemble a salamander—thus coining the term 'Gerry-mander.'

\section{Method}
\begin{figure*}[ht!]
    \centering
    \includegraphics[width=\textwidth]{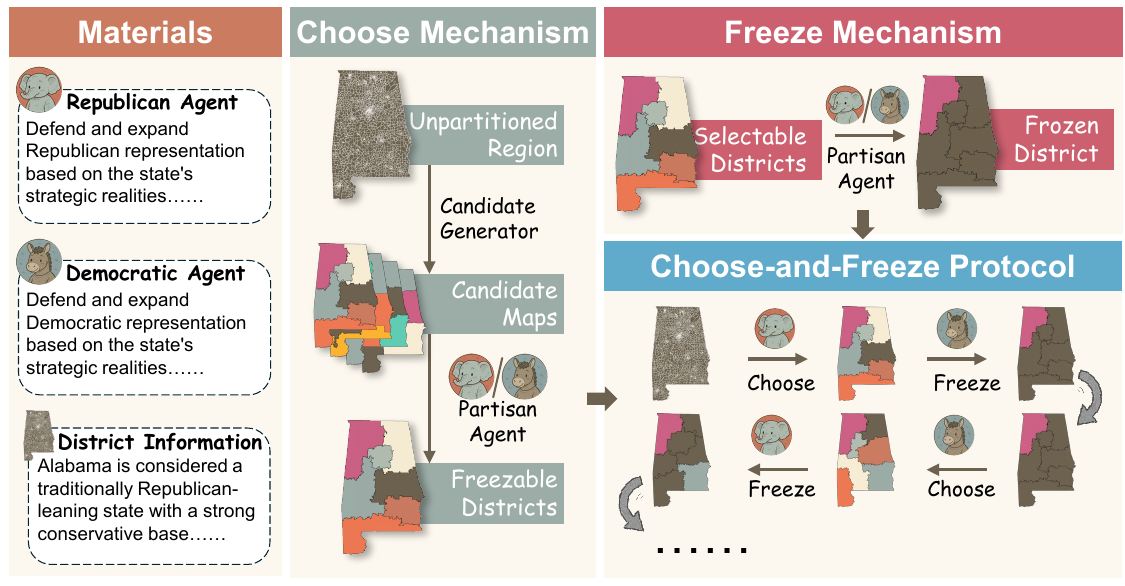}
    \caption{The Agentmandering framework.}
    \label{fig:method}
\end{figure*}

In this section, we provide a detailed introduction to the \textbf{Agentmandering} framework. Agentmandering models redistricting as a structured interaction between two competing agents over a sequence of map construction rounds. As illustrated in Figure~\ref{fig:method}, the method consists of four core components:
(1) Materials: a set of partisan agents representing competing political interests and corresponding district information,
(2) Protocol: a Choose-and-Freeze protocol that alternates these actions until the full state is partitioned,
(3) Choose Mechanism: a candidate generator that proposes feasible districting plans over the current unassigned region, and
(4) Freeze Mechanism: a freeze mechanism that allows the opposing agent to lock in one district per round.

\paragraph{Materials}
The Agentmandering framework operates with two core agents: a Republican agent \(\mathcal{A}_R\) and a Democratic agent \(\mathcal{A}_D\), each representing the strategic interests of one major political party. These agents are powered by LLMs and are prompted to act in alignment with their respective party goals to defend and expand Republican or Democratic representation based on local demographic and political conditions.

Each agent is equipped with a state-specific political profile \(\mathcal{P}_{state}\), which includes historical voting trends, demographic composition (sourced from Census data), and partisan geography. This information, combined with racial demographics, provides strategic cues that guide the agent's behavior throughout the redistricting game.

\paragraph{Choose-and-Freeze Protocol} 
\label{sec:game-loop}
The core of the Agentmandering framework is a sequential game played between a Republican agent \( \mathcal{A}_R \) and a Democratic agent \( \mathcal{A}_D \), who alternate roles over a series of rounds indexed by \( n = 1, 2, \dots, N \). Here \(N\) represents the total number of districts in the corresponding state. At each round, the agents jointly construct the redistricting map by interacting over a progressively shrinking unpartitioned region \( \mathcal{R}_t \). At initialization, the unpartitioned region \( \mathcal{R}_0 \) is the entire state.

Each round consists of two key actions:
\begin{itemize}
    \item \textbf{Choose:} One agent, denoted \( \mathcal{A}_c \in \{\mathcal{A}_R, \mathcal{A}_D\} \), selects a preferred districting plan \( M_n^* \) from a small set of candidate maps \( \mathcal{C}_n \) generated over the current region \( \mathcal{R}_n \).
    \item \textbf{Freeze:} The opposing agent \( \mathcal{A}_{f} \in \{\mathcal{A}_D, \mathcal{A}_R\}\) where \( \mathcal{A}_{f} \neq  \mathcal{A}_c\) selects one district \( D_n^* \in M_n^* \) to be permanently fixed. The remaining territory is updated as \( \mathcal{R}_{n+1} = \mathcal{R}_n \setminus D_n^* \).
\end{itemize}
This iterative process continues until the entire territory has been partitioned into districts. In each round, the choose and freeze agents will be switched. The game structure ensures that no single agent can unilaterally control the full outcome; instead, the final map emerges through a series of constrained, adversarial decisions.

\paragraph{Choose Mechanism}
\label{sec:choose}
In each round \( n \), the candidate generator \( \mathcal{G}(\cdot) \) produces a set of feasible candidate maps \( \mathcal{C}_n \) over the current unpartitioned region \( \mathcal{R}_n \). \( \mathcal{G}(\cdot) \) is party-agnostic and shared by both sides. The size of \( \mathcal{C}_n \) is \( c \), and each plan \( M_i \in \mathcal{C}_n \) satisfies population balance, contiguity, and legal constraints. The choice of \( \mathcal{G}(\cdot) \) is flexible, in this work, we adopt the ReCom~\cite{recom} algorithm. Then the choosing agent \( \mathcal{A}_c \in \{\mathcal{A}_R, \mathcal{A}_D\} \) selects one plan:

\[
M_n^* = f_{\mathrm{choose}}(\mathcal{A}_c, \mathcal{C}_n, \mathcal{P}_{state}),
\]
where \( \mathcal{P}_{state} \) denotes the political profile of the state and \( f_{\mathrm{choose}} \) is a prompt-driven district selection function. The selected map \( M_n^* \) is then passed to the opposing agent for the freeze step.

\paragraph{Freeze Mechanism}
\label{sec:freeze}
Upon receiving the selected plan \( M_n^* \), the freezing agent \( \mathcal{A}_f \), selects one district \( D_n^* \in M_n^* \) to be permanently assigned. The selection is performed via a prompt-driven strategy function:
\[
D_n^* = f_{\mathrm{freeze}}(\mathcal{A}_f, M_n^*, \mathcal{P}_{state}),
\]
where \( f_{\mathrm{freeze}} \) evaluates each district in \( M_n^* \) based on its partisan composition and strategic implications for the freezing agent. Once frozen, the district \( D_n^* \) is removed from the unassigned region:
\[
\mathcal{R}_{n+1} = \mathcal{R}_n \setminus D_n^*.
\]
This procedure ensures that both agents influence the final map through alternating constrained actions, maintaining strategic balance throughout the game. 


\section{Experiments}
\label{sec:experiments}
\subsection{Datasets, Baselines, and Metrics}
We evaluate the effectiveness of the Agentmandering framework using redistricting data from U.S. states based on the post-2020 Census. The data includes population census data, voting history, demographic composition for each state as of 2020, as well as voting data from the 2020 presidential election.\footnote{The district geographic information data is sourced from \url{https://data.census.gov}, and the voting data is from \url{https://dataverse.harvard.edu/dataverse/electionscience}.} The experiments use several baselines including Recom~\cite{recom}, Merge-Split~\cite{Carter2019-ge}, FlipMCMC~\cite{fifield2020automated}, and SMCredist~\cite{McCartan2020-uz}. 


The evaluation metrics used in our experiments include \textbf{Population Deviation (PD)}, \textbf{Polsby-Popper Score (PPS)}, \textbf{Partisan Bias (Bias)}, and \textbf{Local Unfairness (Unfairness)}. 
\textit{Population Deviation} measures the average population imbalance across districts; lower values indicate better population equality. 
\textit{Polsby--Popper Score} evaluates the geometric compactness of each district; higher scores are preferred. 
\textit{Partisan Bias} quantifies the asymmetry in partisan advantage; values closer to zero indicate fairer representation without systematic favor toward either Democrats or Republicans. 
\textit{Local Unfairness} captures the extent to which voter preferences are respected within neighboring districts; lower values imply better local representational equity. 

\subsection{Performance of Agentmandering}
\label{sec:fairness}

\paragraph{Experiment Setup} In this experiment, we evaluate whether Agentmandering produces fairer districting plans compared to traditional ensemble-based sampling methods. Both approaches rely on generating a large number of valid districting plans under legal and demographic constraints, but differ in how these plans are used. We use Gemini 2.5 pro (gemini-2.5-pro-preview-05-06) as our base LLM, and ReCom~\cite{recom} as our Candidate Generator. The temperature was set to 0 for all models and agent steps in every experiment.

In Agentmandering, each round involves selecting one district from a small set of \( c \) candidate maps, and a full run covers \( t \) districts in total. This results in \( l = c \times t \) samples per run. For example, in Alabama, where \( t = 7 \) and \( c = 100 \), one complete Agentmandering game uses around 700 samples. If we repeat this process 10 times, the total number of samples is 7000. 


To ensure a fair comparison, we allow the baseline method to generate the same total number of plans. However, unlike Agentmandering, which incrementally builds maps through an interactive process over a shrinking unassigned region, the baseline produces complete maps in one step. As a result, Agentmandering generates fewer final maps, but each is shaped through strategic agent interactions and controlled partisan dynamics.

\begin{table*}[!htp]
    \centering

    \begin{tabular}{l|cccccc}
    \toprule
    \textbf{Metric} & \textbf{Flip} & \textbf{Merge-split} & \textbf{SMC} & \textbf{Recom} & \textbf{CD-2020} & \textbf{Agentmandering} \\
    \midrule
    \multicolumn{7}{c}{\textit{Arizona (AZ)}} \\ 
    \midrule
    PD(10$^{-3}$)         & 4.59$\pm$1.13 & 4.69$\pm$0.97 & 4.72$\pm$0.91 & 4.72$\pm$1.07 & 62.9 & \textbf{4.19$\pm$0.00324} \\
    PPS(10$^{-2}$)        & 6.75$\pm$2.34 & \textbf{7.02}$\pm$2.36 & 6.97$\pm$2.31 & 6.77$\pm$2.19 & 1.06 & 4.77$\pm$\textbf{0.00317} \\
    Bias(10$^{-2}$)       & \textbf{2.68}$\pm$0.44 & \textbf{2.68}$\pm$0.46 & 2.69$\pm$0.46 & \textbf{2.68}$\pm$0.44 & 3.05 & 3.41$\pm$\textbf{0.00361} \\
    Unfairness(10$^{-1}$) & 4.28$\pm$0.09 & 4.28$\pm$0.08 & 4.28$\pm$0.09 & 4.27$\pm$0.09 & 4.24 & \textbf{4.03$\pm$0.00362} \\
    \midrule
    \multicolumn{7}{c}{\textit{Georgia (GA)}} \\
    \midrule
    PD(10$^{-3}$)         & 4.58$\pm$0.76 & 4.49$\pm$0.74 & \textbf{4.08}$\pm$0.74 & 4.73$\pm$0.78 & 76.6 & 6.15$\pm$\textbf{0.00493} \\
    PPS(10$^{-2}$)        & \textbf{4.68}$\pm$0.68 & 4.62$\pm$0.63 & 4.55$\pm$0.56 & 4.57$\pm$0.69 & 0.75 & 4.04$\pm$\textbf{0.00290} \\
    Bias(10$^{-3}$)       & 8.03$\pm$2.19 & 8.12$\pm$2.33 & 8.15$\pm$1.56 & 8.13$\pm$2.38 & 6.53 & \textbf{7.63$\pm$0.00508} \\
    Unfairness(10$^{-1}$) & 3.68$\pm$0.07 & 3.68$\pm$0.07 & 3.66$\pm$0.05 & 3.67$\pm$0.07 & 3.56 & \textbf{3.49$\pm$0.00333} \\
    \midrule
    \multicolumn{7}{c}{\textit{Michigan (MI)}} \\
    \midrule
    PD(10$^{-3}$)         & 4.70$\pm$0.76  & 4.72$\pm$0.70  & 4.72$\pm$0.72  & 4.31$\pm$0.83  & 95.2  & \textbf{3.70$\pm$0.00237}  \\
    PPS(10$^{-2}$)        & \textbf{6.03}$\pm$0.81  & 5.84$\pm$0.87  & 5.92$\pm$0.87  & 6.00$\pm$0.84  & 0.74  & 5.94$\pm$\textbf{0.00429}  \\
    Bias(10$^{-2}$)       & -2.34$\pm$0.14 & -2.34$\pm$0.14 & -2.34$\pm$0.13 & \textbf{-2.30}$\pm$0.14 & -2.06 & -2.40$\pm$\textbf{0.00415} \\
    Unfairness(10$^{-1}$) & 4.20$\pm$0.08  & 4.18$\pm$0.07  & 4.18$\pm$0.08  & 4.19$\pm$0.08  & 4.10  & \textbf{3.96$\pm$0.00280}  \\
    \midrule
    \multicolumn{7}{c}{\textit{North Carolina (NC)}} \\
    \midrule
    PD(10$^{-3}$)         & 4.73$\pm$0.73  & 4.61$\pm$0.77  & 4.64$\pm$0.74  & 4.57$\pm$0.75  & 94.2  & \textbf{3.72$\pm$0.00196}  \\
    PPS(10$^{-2}$)        & 5.98$\pm$0.81  & 6.00$\pm$0.88  & \textbf{6.11}$\pm$0.97  & 5.87$\pm$0.91  & 0.85  & 5.31$\pm$\textbf{0.00186}  \\
    Bias(10$^{-2}$)       & -2.36$\pm$0.14 & -2.37$\pm$0.15 & -2.37$\pm$0.15 & -2.32$\pm$0.15 & -2.07 & \textbf{-2.22$\pm$0.00134} \\
    Unfairness(10$^{-1}$) & 4.18$\pm$0.09  & 4.19$\pm$0.09  & 4.19$\pm$0.08  & 4.19$\pm$0.09  & 4.15  & \textbf{3.94$\pm$0.00209}  \\
    \midrule
    \multicolumn{7}{c}{\textit{Nevada (NV)}} \\
    \midrule
    PD(10$^{-3}$)         & 4.20$\pm$1.44 & 4.30$\pm$1.43 & 4.06$\pm$1.47 & 4.31$\pm$1.44 & 49.3 & \textbf{4.03$\pm$0.00292} \\
    PPS(10$^{-1}$)        & 2.04$\pm$0.35 & 2.08$\pm$0.39 & \textbf{2.12}$\pm$0.38 & 2.07$\pm$0.37 & 0.32 & 2.05$\pm$\textbf{0.00212} \\
    Bias(10$^{-2}$)       & 4.91$\pm$0.57 & 4.93$\pm$0.54 & 4.95$\pm$0.54 & 4.87$\pm$0.54 & 3.17 & \textbf{4.57$\pm$0.00419} \\
    Unfairness(10$^{-1}$) & 4.39$\pm$0.04 & 4.39$\pm$0.03 & 4.39$\pm$0.03 & 4.39$\pm$0.03 & 4.29 & \textbf{4.30$\pm$0.00468} \\
    \midrule
    \multicolumn{7}{c}{\textit{Pennsylvania (PA)}} \\
    \midrule
    PD(10$^{-3}$)         & 4.63$\pm$0.73 & \textbf{4.32}$\pm$0.86 & 4.42$\pm$0.75 & 4.39$\pm$0.78 & 73.5 & 4.88\textbf{$\pm$0.00297} \\
    PPS(10$^{-2}$)        & \textbf{4.84}$\pm$0.74 & 4.72$\pm$0.69 & 4.70$\pm$0.63 & 4.63$\pm$0.63 & 0.76 & 4.19$\pm$\textbf{0.00323} \\
    Bias(10$^{-3}$)       & 8.08$\pm$2.43 & 8.58$\pm$2.29 & 7.83$\pm$2.47 & 8.49$\pm$2.32 & 6.59 & \textbf{7.86$\pm$0.00550} \\
    Unfairness(10$^{-1}$) & 3.68$\pm$0.07 & 3.68$\pm$0.07 & 3.69$\pm$0.07 & 3.66$\pm$0.07 & 3.61 & \textbf{3.48$\pm$0.00315} \\
    \midrule
    \multicolumn{7}{c}{\textit{Wisconsin (WI)}} \\
    \midrule
    PD(10$^{-3}$)         & 4.13$\pm$0.98  & 4.31$\pm$1.01  & 4.25$\pm$0.95  & 4.25$\pm$0.97  & 30.5 & \textbf{3.54$\pm$0.00264} \\
    PPS(10$^{-2}$)        & 9.39$\pm$1.65  & 9.39$\pm$1.65  & 9.43$\pm$1.74  & 9.48$\pm$1.68  & 1.22 & \textbf{9.56$\pm$0.00384} \\
    Bias(10$^{-3}$)       & -1.20$\pm$2.57 & -1.39$\pm$2.29 & -1.11$\pm$2.31 & -1.32$\pm$2.35 & 8.08 & \textit{1.29}$\pm$\textbf{0.00063} \\
    Unfairness(10$^{-1}$) & 4.20$\pm$0.11  & 4.20$\pm$0.11  & 4.21$\pm$0.11  & 4.21$\pm$0.11  & 3.96 & \textbf{3.93$\pm$0.00271} \\
    \bottomrule
    \end{tabular}
    \caption{Experimental Results on Key Swing States: Bold indicates the best performance (with mean and standard deviation separated), and italics denote the same bias as the real situation, reflected in the sign of the Bias metric.}
    
    \label{tab:performance-results}
    \end{table*}

\begin{table*}[!htp]
    \centering

    \setlength{\tabcolsep}{3pt}  
    \renewcommand{\arraystretch}{0.9}  

    \begin{tabular}{l|ccccccccc}
    \toprule
    \textbf{State} & Gemini & GPT-4om & GPT-o3m & DS-R1 & DS-V3 & Claude-3.7 & Llama3 & Mixtral3.1 & Qwen3 \\
    \midrule
    AZ & 4.03$_{3.61}$ & 4.10$_{2.71}$ & 4.07$_{3.33}$ & 4.04$_{2.57}$ & 4.05$_{2.40}$ & 4.04$_{3.91}$ & 4.06$_{3.37}$ & 4.08$_{4.40}$ & 4.10$_{4.89}$ \\
    GA & 3.49$_{3.33}$ & 3.48$_{2.28}$ & 3.48$_{2.43}$ & 3.48$_{1.51}$ & 3.48$_{3.15}$ & 3.49$_{4.53}$ & 3.50$_{2.42}$ & 3.48$_{4.45}$ & 3.48$_{5.52}$ \\
    MI & 3.96$_{2.80}$ & 3.95$_{2.85}$ & 3.95$_{2.60}$ & 3.96$_{2.48}$ & 3.95$_{2.19}$ & 3.97$_{5.08}$ & 3.97$_{1.66}$ & 3.97$_{5.32}$ & 3.95$_{4.61}$ \\
    NC & 3.94$_{2.09}$ & 3.98$_{2.27}$ & 3.96$_{2.03}$ & 4.04$_{2.05}$ & 4.05$_{2.80}$ & 3.97$_{3.99}$ & 4.06$_{2.02}$ & 3.94$_{4.67}$ & 3.98$_{3.30}$ \\
    NV & 4.30$_{4.68}$ & 4.29$_{4.68}$ & 4.38$_{3.36}$ & 4.34$_{3.75}$ & 4.26$_{4.56}$ & 4.37$_{5.92}$ & 4.36$_{4.12}$ & 4.30$_{3.97}$ & 4.29$_{4.80}$ \\
    PA & 3.48$_{3.15}$ & 3.48$_{2.69}$ & 3.47$_{3.25}$ & 3.47$_{2.99}$ & 3.48$_{3.89}$ & 3.46$_{2.79}$ & 3.48$_{3.58}$ & 3.49$_{3.96}$ & 3.49$_{3.53}$ \\
    WI & 3.93$_{2.71}$ & 3.97$_{2.43}$ & 3.95$_{2.61}$ & 3.94$_{2.44}$ & 3.96$_{2.14}$ & 3.96$_{3.99}$ & 3.94$_{2.29}$ & 3.96$_{4.47}$ & 3.97$_{5.11}$ \\
    \bottomrule
    \end{tabular}
    \caption{Performance of Agentmandering across states using different LLMs. The mean values are in scientific notation (10$^{-1}$), and the standard deviations are in scientific notation (10$^{-4}$).}
    \label{tab:llm-comparison}
    \end{table*}
\paragraph{Results}
Table~\ref{tab:performance-results} reports results on seven competitive swing states, with bold text indicating the best scores. The row \textbf{CD\_2020} represents enacted districting plans. Agentmandering shows a strong advantage in \emph{stability}, with standard deviations at least two orders of magnitude smaller than other methods, indicating reduced metric fluctuation and less room for strategic manipulation.

On \textbf{PD} (Population Deviation) and \textbf{PPS} (Polsby–Popper Score), all computational methods outperform CD\_2020, revealing population imbalance and geometric distortion in real-world plans. Agentmandering slightly underperforms on PPS due to its irregular boundaries but achieves the lowest PD, reflecting superior population balance.

For \textbf{Bias} and \textbf{Unfairness}, which assess partisan neutrality and representational equity, Agentmandering performs best or near-best across most states. Notably, it is the only method to recover the correct partisan direction in Wisconsin, highlighted in italics. Its low Unfairness scores suggest more balanced and satisfying outcomes for voters.

\begin{figure*}[ht!]
\centering
\includegraphics[width=0.9\textwidth]{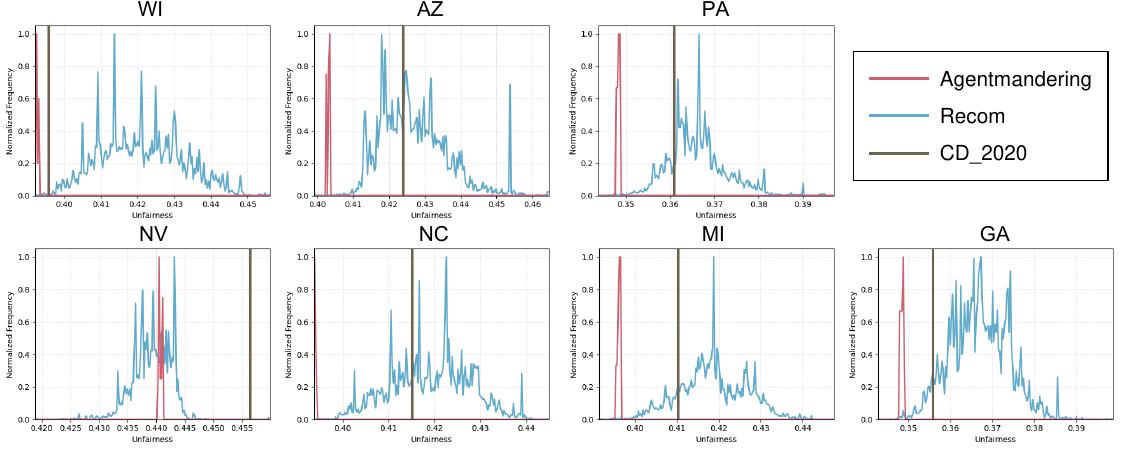}
\caption{Normalized distribution of Unfairness across seven states for Agentmandering and Recom.}
\label{fig:dis}
\end{figure*}

Figure~\ref{fig:dis} shows that Agentmandering produces fairer and more stable maps than both Recom and CD\_2020, despite using Recom for candidate generation. This demonstrates its robustness and capacity to mitigate partisan bias.

\subsection{Effectiveness of LLM-Based Agent Decisions}

\paragraph{Experiment Setup}  
In this section, we evaluate the effectiveness of LLM-based agents' decisions within the Agentmandering framework.  
As a baseline, we implement rule-based variants for both the \textit{choose} and \textit{freeze} steps.  
The evaluation metric used is \textbf{Unfairness}.  
Specifically, we compare against the following decision rules:

\begin{itemize}
    \item \textbf{Partisan Bias:} In the \textit{choose} step, the agent selects the map that maximizes partisan advantage for its affiliated party; in the \textit{freeze} step, it freezes the district that offers the greatest partisan gain.
    \item \textbf{Population Deviation:} In both \textit{choose} and \textit{freeze} steps, the agent selects the map or district with the smallest population deviation.
    \item \textbf{Compactness:} In both steps, the agent chooses the most compact option according to the Polsby–Popper score.
\end{itemize}

\begin{figure*}[ht!]
    \centering
    \includegraphics[width=0.85\textwidth]{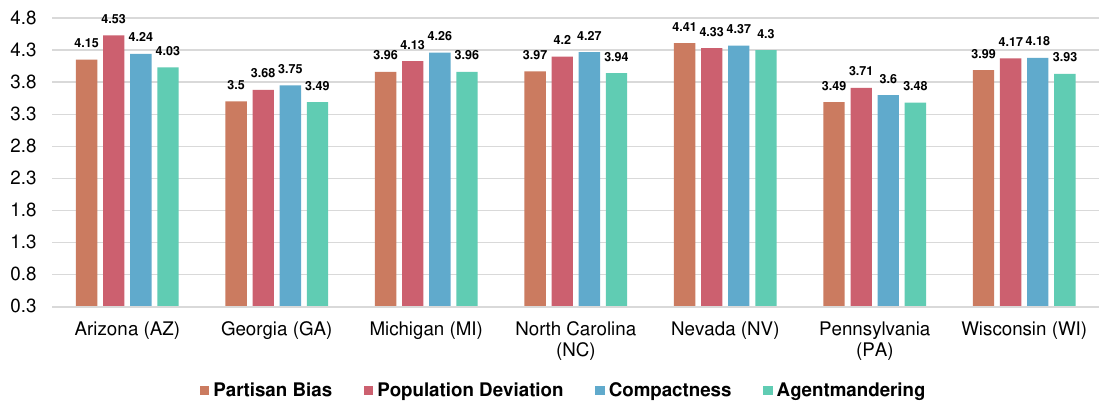}
    \caption{Comparison of LLM-based and rule-based agent decisions in terms of Unfairness.}
    \label{fig:rule}
\end{figure*}

\paragraph{Results}  
The results are shown in Figure~\ref{fig:rule}.  
As observed, \textbf{Agentmandering achieves a lower Unfairness score than all rule-based variants}.  
This indicates that LLM-based agents are more effective in making politically strategic decisions—both in selecting appropriate candidate maps and in freezing reasonable districts.  
These results provide strong support for the value of integrating LLM reasoning with a game-theoretic mechanism to simulate human-like political behavior.  
Furthermore, the standard deviation of the rule-based variants is also 2–3 orders of magnitude lower than traditional baselines, suggesting that the overall stability of Agentmandering primarily stems from the \textit{Choose-and-Freeze} protocol.

\subsection{Effect of LLM Choice}
\label{sec:llm-impact}
We examine whether the choice of LLM affects Agentmandering’s performance, given that redistricting involves politically sensitive reasoning and LLMs may differ in bias~\cite{motoki2024more, rozado2024political}. We evaluate performance using \textbf{Unfairness} across a range of models, including Gemini 2.5 Pro, GPT-4o-mini, GPT-o3-mini, Deepseek-R1, Deepseek-V3, Claude-3.7, LLaMA-3-70B, Mixtral-3.1, and Qwen. The proprietary models are accessed via API; open source models run on a 4×A6000 Linux server.

\paragraph{Results.}
Table~\ref{tab:llm-comparison} shows that all models, including those developed in the United States (such as OpenAI and Anthropic), Europe (such as Mistral), and China (such as Deepseek and Qwen), achieve similar Unfairness scores across states and configurations. This consistency across national and institutional contexts suggests that Agentmandering is robust to differences in model origin, training data, or political orientation. The \textit{Choose-and-Freeze} strategy provides sufficient structural guidance to ensure fairness, even when underlying LLMs vary, enabling institutions to substitute or upgrade models without degrading performance.

\section{Conclusion}

We introduce \textbf{Agentmandering}, a novel redistricting framework that harnesses large language model (LLM) agents to implement game-theoretic negotiation in practice. By simulating the \textit{Choose-and-Freeze} protocol through interactive LLM agents, our approach transforms an abstract fairness mechanism into a scalable solution for real-world redistricting challenges. The resulting plans are procedurally transparent, strategically robust, and empirically fair across multiple metrics. This work makes two key contributions. First, it provides a new computational lens for political science by demonstrating how LLMs can model strategic partisan behavior in institutional settings. Second, it shows how LLM agents can bridge game-theoretic fairness and applied algorithmic decision-making.

Looking forward, future work will extend to investigate the challenges of applying this framework to multi-party systems, where fairness and strategy must be redefined to accommodate diverse party dynamics, coalition effects, and proportionality requirements.

\appendix

\section{A1. Prompt Design}
\label{app:prompt}

\paragraph{State Background Prompt.}
The state background prompt provides detailed contextual information to support redistricting decisions. It outlines the state's political identity, partisan geography (urban vs. rural divides, suburban trends), demographic composition (racial breakdown, minority concentrations), electoral history (presidential, gubernatorial, legislative trends), and strategic considerations for both parties. This structured information equips agents with the political and demographic landscape necessary to make informed and realistic districting decisions.

\paragraph{Agent Prompt.}
The agent prompt defines the role of a partisan negotiator—either Democratic or Republican—tasked with maximizing their party's electoral advantage while complying with legal constraints. It includes state-type-specific strategies (red, swing, blue states), core negotiation principles (e.g., packing, cracking, defending urban cores), and a fixed evaluation format. The agent must select one district to freeze in every round, and justify the choice in terms of long-term party interest, even under suboptimal conditions. This prompt enables LLM agents to simulate realistic, asymmetric political behavior grounded in strategic reasoning.

\section{A2. Additional Experimental Settings}
\label{sec:additional-settings}
\paragraph{Baselines.}
We compare Agentmandering against four representative sampling-based redistricting algorithms.

\begin{itemize}
    \item \textbf{Flip}~\cite{fifield2020automated}: A classical MCMC method that performs local moves by reassigning a single boundary node to a neighboring district while preserving contiguity, offering simplicity but limited mixing efficiency in high-constraint settingsReCom.
    \item \textbf{ReCom}~\cite{recom}: Recombination improves mixing by merging two adjacent districts and repartitioning them via spanning trees, producing globally structured changes in district mapsReCom.
    \item \textbf{Merge-Split}~\cite{Carter2019-ge}: Merge-Split extends ReCom by using reversible spanning-tree-based proposals in Metropolis-Hastings sampling, enabling efficient exploration of large redistricting spaces with guaranteed detailed balancemergesplit.
    \item \textbf{SMC}~\cite{McCartan2020-uz}: SMC (Sequential Monte Carlo) draws redistricting plans iteratively from scratch using particle filters, allowing scalable, parallelizable sampling that accommodates real-world constraints more flexibly than traditional MCMC methods.
\end{itemize}

\section{A3. Additional Experimental Results}

\subsection{Performance Results on Additional States}
\label{sec:additional-performance-results}
\begin{figure*}[ht!]
    \centering
    \includegraphics[width=\textwidth]{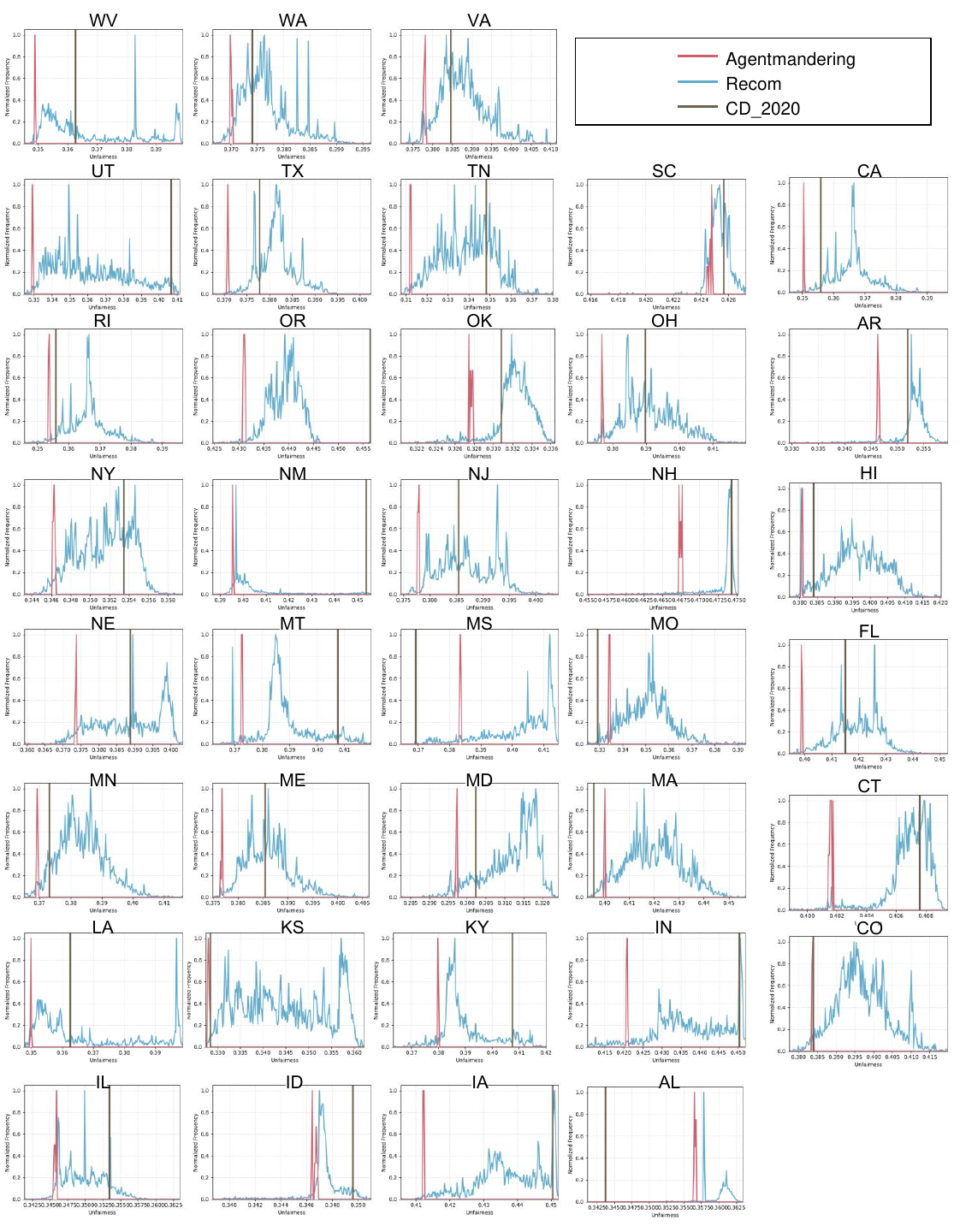}
    \caption{Distribution of \textit{Unfairness} scores across additional U.S. states, comparing Agentmandering and Recom. Lower scores indicate better voter satisfaction.}
    \label{fig:appdis}
\end{figure*}

Figures~\ref{fig:appdis} visualize the distribution of \textit{Unfairness} scores produced by Agentmandering and Recom across additional U.S. states. The results demonstrate that Agentmandering consistently generates districting plans that are more compact and fairer than those produced by baseline methods. Although in a few states Agentmandering underperforms compared to the enacted plans (CD\_2020), the overall trend confirms the framework's strong ability to produce high-quality, procedurally balanced districting outcomes.

\subsection{Effect of Repeated Runs}
\label{app:repeat-runs}

To evaluate the stability of Agentmandering under repeated execution, we conduct an experiment on Pennsylvania by varying the number of independent runs. While our main experiments use 10 repeated runs per setting, reviewers may be concerned that such limited sampling could mask instability. To address this, we execute Agentmandering with 10, 50, 100, 200, and 500 independent runs and evaluate the variance and mean of key fairness metrics.

The results, summarized in Table~\ref{tab:repeats}, show that both the mean and standard deviation of all metrics remain remarkably stable as the number of runs increases. This supports our claim that Agentmandering produces robust and reproducible outcomes even with moderate sampling, and that its performance does not degrade under more extensive evaluation.

\begin{table*}[ht]
    \centering
    \small
    \caption{Stability of Agentmandering across different numbers of repeated runs on Pennsylvania. Values are shown as mean ± standard deviation. PD is scaled by \( \times 10^{-3} \), PPS by \( \times 10^{-2} \), Bias by \( \times 10^{-3} \), and Unfairness by \( \times 10^{-1} \).}
    \label{tab:repeats}
    \begin{tabular}{l|ccccc}
        \toprule
        Metric & 10 Runs & 50 Runs & 100 Runs & 200 Runs & 500 Runs \\
        \midrule
        PD         & 4.84 ± 0.0023 & 4.11 ± 0.0017 & 4.49 ± 0.0029 & 4.71 ± 0.0031 & 4.53 ± 0.0023 \\
        PPS        & 3.13 ± 0.0020 & 3.01 ± 0.0017 & 2.98 ± 0.0015 & 2.95 ± 0.0015 & 3.00 ± 0.0014 \\
        Bias       & 5.05 ± 0.0031 & 4.58 ± 0.0019 & 4.78 ± 0.0023 & 4.65 ± 0.0025 & 4.88 ± 0.0019 \\
        Unfairness & 3.47 ± 0.0019 & 3.47 ± 0.0021 & 3.46 ± 0.0014 & 3.48 ± 0.0016 & 3.47 ± 0.0016 \\
        \bottomrule
    \end{tabular}
\end{table*}

These findings reinforce the methodological soundness of our evaluation: even with limited runs, Agentmandering consistently converges to stable outcomes, validating the reliability of our fairness assessments.

\subsection{Running Time}
\label{app:time}
We record the running time of Agentmandering using Gemini 2.5 Pro as the LLM agent and Recom as the candidate generator, with 100 candidate maps generated in each \textit{choose} step. Table~\ref{tab:runtime} reports the average runtime per state in seconds. The number of districts in each state is indicated in parentheses.
\begin{table*}[ht]
\centering
\caption{Average running time (in seconds) of Agentmandering in each state using Gemini 2.5 Pro and Recom with 100 candidate maps.}
\label{tab:runtime}
\begin{tabular}{lcccccccccc}
\toprule
State & AK(1) & AL(7) & AR(4) & AZ(9) & CA(52) & CO(8) & CT(5) & DE(1) & FL(28) & GA(14) \\
Time (s) & - & 14.5 & 10.94 & 13.28 & 124.43 & 14.66 & 9.97 & - & 43.14 & 22.32 \\
\midrule
State & HI(2) & IA(4) & ID(2) & IL(17) & IN(9) & KS(4) & KY(6) & LA(6) & MA(9) & MD(8) \\
Time (s) & 8.52 & 10.73 & 8.22 & 30.35 & 15.23 & 11.02 & 14.64 & 14.93 & 14.04 & 14.72 \\
\midrule
State & ME(2) & MI(13) & MN(8) & MO(8) & MS(4) & MT(2) & NC(14) & ND(1) & NE(3) & NH(2) \\
Time (s) & 8.47 & 21.09 & 13.88 & 14.73 & 10.72 & 9.82 & 21.69 & - & 10.88 & 9.45 \\
\midrule
State & NJ(12) & NM(3) & NV(4) & NY(26) & OH(15) & OK(5) & OR(6) & PA(17) & RI(2) & SC(7) \\
Time (s) & 23.85 & 11.41 & 10.11 & 41.59 & 22.06 & 10.76 & 15.22 & 32.76 & 9.14 & 14.05 \\
\midrule
State & SD(1) & TN(9) & TX(38) & UT(4) & VA(11) & VT(1) & WA(10) & WI(8) & WV(2) & WY(1) \\
Time (s) & - & 13.84 & 81.88 & 9.53 & 21.22 & - & 14.77 & 14.47 & 9.25 & - \\
\bottomrule
\end{tabular}
\end{table*}

As shown in Table~\ref{tab:runtime}, runtime increases with the number of districts. States with only one district are excluded, as no redistricting is required. Most full Agentmandering runs complete in under 30 seconds, demonstrating the method’s practical efficiency and suitability for large-scale deployment across all U.S. states.

\subsection{Additional Ablation Study}
\label{app:ablation-study}
In this section, we investigate three key factors that may affect the performance of Agentmandering:

\paragraph{Candidate set size for the \textit{choose} step}: We experiment with five different sizes of candidate maps—10, 50, 100, 200, and 500—to assess how the size of the selection pool impacts outcomes.

We examine how the number of candidate maps available during each \textit{choose} step influences final outcomes. Specifically, we vary the candidate set size across five levels: 10, 50, 100, 200, and 500. As shown in Table~\ref{tab:ablation-study-results}, increasing the size of the candidate pool consistently improves performance across multiple fairness metrics, including population deviation, partisan bias, and compactness. This trend reflects the intuitive advantage of offering more diverse options for strategic selection. However, we observe diminishing returns beyond a candidate size of 100, suggesting a balance between computational cost and fairness gain. Consequently, we set the default candidate size to 100 in all main experiments to ensure both efficiency and effectiveness.

\paragraph{Agent ordering}: We explore whether the initial \textit{choose} agent influences results by testing two configurations: one where the Republican agent chooses first, and another where the Democratic agent does.

We also evaluate whether the identity of the first agent to act (i.e., the agent that performs the initial \textit{choose} step) affects overall outcomes. To this end, we conduct two runs of Agentmandering for each state: one where the Republican agent moves first, and one where the Democratic agent does. The results indicate that the choice of initial agent leads to only marginal variation in fairness metrics, with no systematic advantage observed. This suggests that the alternating structure of the \textit{Choose-and-Freeze} protocol balances power across rounds, making the overall process robust to agent ordering. Such symmetry further highlights the procedural fairness of our framework and supports its generalizability.

\paragraph{Candidate generator \( \mathcal{G}(\cdot) \)}: We compare the impact of two widely used map generation algorithms, Recom and Flip, on the final results produced by Agentmandering.

\begin{figure*}[ht!]
        \centering
        \includegraphics[width=0.9\textwidth]{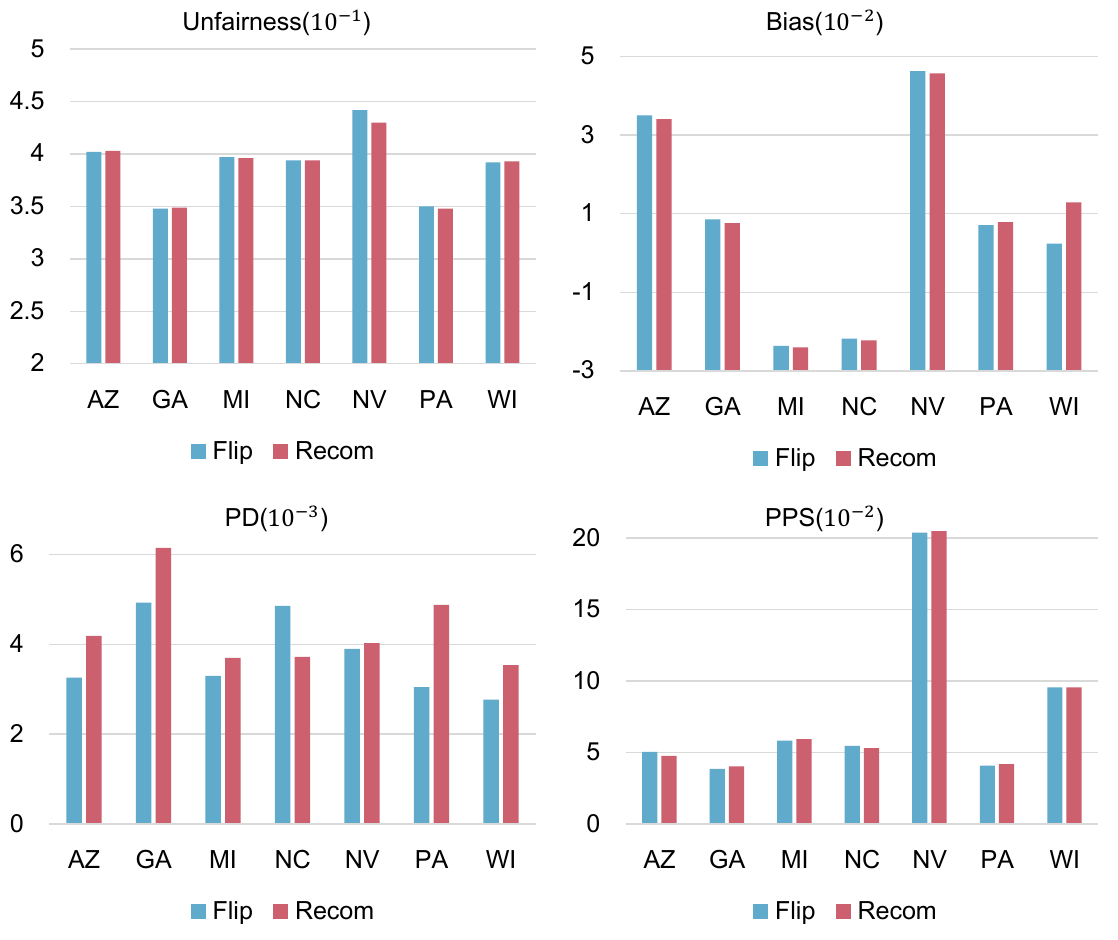}
        \caption{Comparison of Agentmandering performance using two candidate generators: Recom and Flip.}
        \label{fig:flip}
    \end{figure*}

Results of the third experiment are shown in Figure~\ref{fig:flip}. We compare two different candidate generators—Recom and Flip—within the Agentmandering framework to evaluate their impact on overall performance. The results indicate that both generators perform similarly on \textit{Partisan Bias}, \textit{Unfairness}, and \textit{Polsby–Popper Score}, suggesting that our method is robust to the choice of generator. This reflects the flexibility of the Choose-and-Freeze protocol, which can accommodate a wide range of districting functions. On the \textit{Population Deviation} metric, Flip performs slightly better, indicating its stronger ability to produce plans with more balanced population distributions.

\begin{table*}[ht]
    \centering
    \caption{Ablation study results on key swing states. The notation "d10" indicates that 10 candidate maps are generated per round and that the Democratic agent chooses first. All metric values are reported in scientific notation: PD mean is scaled by \( \times 10^{-3} \), standard deviation by \( \times 10^{-6} \); PPS mean by \( \times 10^{-2} \), standard deviation by \( \times 10^{-5} \); Bias mean by \( \times 10^{-2} \), standard deviation by \( \times 10^{-5} \); and Unfairness mean by \( \times 10^{-1} \), standard deviation by \( \times 10^{-4} \).}
    \label{tab:ablation-study-results}
    \renewcommand{\arraystretch}{0.8}
    \begin{tabular}{l|cccccccccc}
    \toprule
    \textbf{Metric} & \textbf{d10} & \textbf{d50} & \textbf{d100} & \textbf{d200} & \textbf{d500} & \textbf{r10} & \textbf{r50} & \textbf{r100} & \textbf{r200} & \textbf{r500} \\
    \midrule
    \multicolumn{11}{c}{\textit{Arizona (AZ)}} \\
    \midrule
    PD & 4.49$_{4.58}$ & 4.34$_{2.87}$ & 4.31$_{4.93}$ & 4.62$_{4.79}$ & 4.53$_{2.37}$ & 4.20$_{5.16}$ & 4.16$_{4.82}$ & 4.36$_{4.35}$ & 4.42$_{4.14}$ & 4.09$_{5.73}$ \\
    PPS & 8.45$_{8.62}$ & 4.92$_{6.37}$ & 4.79$_{4.77}$ & 4.70$_{5.27}$ & 4.80$_{5.20}$ & 8.08$_{9.07}$ & 4.81$_{5.51}$ & 4.77$_{4.68}$ & 4.79$_{5.92}$ & 4.77$_{5.45}$ \\
    Bias & 5.80$_{5.08}$ & 3.41$_{2.57}$ & 3.42$_{4.49}$ & 3.30$_{3.24}$ & 3.39$_{4.10}$ & 4.87$_{5.94}$ & 3.44$_{4.09}$ & 3.43$_{2.98}$ & 3.41$_{4.75}$ & 3.41$_{3.64}$ \\
    Unfairness & 5.02$_{5.86}$ & 4.03$_{4.11}$ & 4.03$_{3.79}$ & 4.02$_{5.30}$ & 4.03$_{4.16}$ & 4.95$_{5.02}$ & 4.03$_{4.01}$ & 4.03$_{4.18}$ & 4.03$_{5.01}$ & 4.03$_{4.91}$ \\
    \midrule
    \multicolumn{11}{c}{\textit{Georgia (GA)}} \\
    \midrule
    PD & 4.28$_{5.82}$ & 6.11$_{7.29}$ & 6.07$_{7.80}$ & 6.46$_{7.49}$ & 6.20$_{6.34}$ & 3.66$_{3.61}$ & 5.39$_{6.68}$ & 6.16$_{6.72}$ & 6.27$_{7.55}$ & 5.76$_{7.05}$ \\
    PPS & 7.90$_{8.14}$ & 4.01$_{4.19}$ & 4.04$_{5.95}$ & 3.96$_{4.53}$ & 4.09$_{4.15}$ & 7.33$_{9.26}$ & 4.10$_{5.67}$ & 4.04$_{4.77}$ & 4.01$_{4.37}$ & 4.06$_{2.91}$ \\
    Bias & 4.83$_{4.89}$ & 9.16$_{1.02}$ & 7.45$_{8.06}$ & 8.06$_{8.55}$ & 7.19$_{7.95}$ & 4.20$_{4.03}$ & 7.72$_{8.53}$ & 7.44$_{8.21}$ & 6.93$_{8.00}$ & 7.81$_{8.95}$ \\
    Unfairness & 4.34$_{3.37}$ & 3.49$_{5.26}$ & 3.49$_{3.77}$ & 3.49$_{4.13}$ & 3.48$_{4.35}$ & 4.26$_{3.32}$ & 3.49$_{3.61}$ & 3.49$_{4.03}$ & 3.49$_{3.89}$ & 3.50$_{4.00}$ \\
    \midrule
    \multicolumn{11}{c}{\textit{Michigan (MI)}} \\
    \midrule
    PD & 4.33$_{5.62}$ & 4.66$_{6.54}$ & 3.82$_{3.40}$ & 4.04$_{4.76}$ & 3.42$_{3.69}$ & 3.83$_{4.75}$ & 4.12$_{3.49}$ & 3.95$_{3.37}$ & 3.83$_{3.39}$ & 3.89$_{4.71}$ \\
    PPS & 1.01$_{7.45}$ & 5.96$_{6.96}$ & 5.96$_{5.29}$ & 5.93$_{6.64}$ & 5.96$_{5.98}$ & 9.35$_{7.64}$ & 6.00$_{5.09}$ & 5.89$_{5.54}$ & 5.90$_{6.68}$ & 5.90$_{7.00}$ \\
    Bias & 1.33$_{1.37}$ & 2.45$_{2.82}$ & 2.39$_{3.02}$ & 2.43$_{3.02}$ & 2.43$_{2.33}$ & 1.46$_{2.36}$ & 2.43$_{2.54}$ & 2.38$_{2.75}$ & 2.47$_{2.76}$ & 2.34$_{2.85}$ \\
    Unfairness & 4.82$_{5.26}$ & 3.97$_{3.85}$ & 3.96$_{4.55}$ & 3.97$_{4.28}$ & 3.96$_{3.20}$ & 4.80$_{5.89}$ & 3.96$_{3.90}$ & 3.96$_{3.69}$ & 3.97$_{2.59}$ & 3.96$_{5.49}$ \\
    \midrule
    \multicolumn{11}{c}{\textit{North Carolina (NC)}} \\
    \midrule
    PD & 4.84$_{4.12}$ & 4.37$_{3.97}$ & 3.81$_{4.91}$ & 3.01$_{2.58}$ & 3.47$_{3.69}$ & 3.91$_{4.04}$ & 3.43$_{4.48}$ & 3.62$_{4.02}$ & 4.60$_{5.63}$ & 4.38$_{3.99}$ \\
    PPS & 9.63$_{7.91}$ & 5.46$_{5.87}$ & 5.33$_{6.41}$ & 5.27$_{4.61}$ & 5.43$_{4.58}$ & 8.69$_{9.64}$ & 5.14$_{5.76}$ & 5.28$_{4.81}$ & 5.28$_{4.76}$ & 5.34$_{5.73}$ \\
    Bias & 2.12$_{1.73}$ & 2.31$_{2.69}$ & 2.22$_{3.19}$ & 2.26$_{3.24}$ & 2.22$_{2.67}$ & 1.25$_{1.57}$ & 2.14$_{2.59}$ & 2.21$_{2.43}$ & 2.24$_{3.26}$ & 2.19$_{2.14}$ \\
    Unfairness & 4.88$_{5.20}$ & 3.95$_{3.96}$ & 3.94$_{4.53}$ & 3.95$_{6.00}$ & 3.94$_{4.82}$ & 4.87$_{5.55}$ & 3.95$_{2.91}$ & 3.95$_{4.50}$ & 3.95$_{5.63}$ & 3.94$_{1.90}$ \\
    \midrule
    \multicolumn{11}{c}{\textit{Nevada (NV)}} \\ 
    \midrule
    PD & 4.36$_{4.35}$ & 4.86$_{5.64}$ & 4.18$_{4.88}$ & 4.56$_{4.27}$ & 3.95$_{4.26}$ & 3.76$_{3.50}$ & 4.29$_{4.90}$ & 4.55$_{5.85}$ & 4.43$_{5.68}$ & 3.66$_{3.80}$ \\
    PPS & 1.52$_{1.31}$ & 2.04$_{2.11}$ & 2.05$_{2.19}$ & 2.05$_{2.20}$ & 2.06$_{2.16}$ & 1.45$_{1.96}$ & 2.05$_{2.39}$ & 2.05$_{2.13}$ & 2.06$_{2.44}$ & 2.05$_{2.53}$ \\
    Bias & 8.13$_{7.65}$ & 4.65$_{5.56}$ & 4.56$_{4.52}$ & 4.60$_{5.51}$ & 4.57$_{5.91}$ & 7.52$_{6.57}$ & 4.47$_{4.83}$ & 4.55$_{5.24}$ & 4.57$_{5.09}$ & 4.58$_{3.43}$ \\
    Unfairness & 4.94$_{5.33}$ & 4.41$_{4.73}$ & 4.41$_{3.82}$ & 4.41$_{4.94}$ & 4.41$_{5.69}$ & 4.88$_{5.78}$ & 4.41$_{4.72}$ & 4.41$_{5.15}$ & 4.41$_{4.40}$ & 4.41$_{3.97}$ \\
    \midrule
    \multicolumn{11}{c}{\textit{Pennsylvania (PA)}} \\
    \midrule
    PD & 4.15$_{4.08}$ & 5.12$_{7.31}$ & 4.94$_{4.91}$ & 3.94$_{4.95}$ & 4.61$_{5.72}$ & 3.53$_{4.11}$ & 4.13$_{4.69}$ & 5.22$_{4.82}$ & 4.78$_{6.09}$ & 4.99$_{3.84}$ \\
    PPS & 8.81$_{1.07}$ & 4.13$_{4.98}$ & 4.18$_{5.06}$ & 4.21$_{3.45}$ & 4.17$_{4.00}$ & 8.06$_{8.46}$ & 4.32$_{6.00}$ & 4.18$_{4.98}$ & 4.18$_{3.52}$ & 4.25$_{4.85}$ \\
    Bias & 4.49$_{3.82}$ & 8.43$_{8.35}$ & 7.76$_{7.75}$ & 7.86$_{8.94}$ & 8.14$_{8.55}$ & 4.16$_{3.48}$ & 8.08$_{1.01}$ & 8.23$_{7.89}$ & 7.67$_{9.07}$ & 7.61$_{8.83}$ \\
    Unfairness & 4.29$_{5.42}$ & 3.49$_{2.89}$ & 3.48$_{3.39}$ & 3.48$_{2.54}$ & 3.48$_{3.68}$ & 4.26$_{3.57}$ & 3.48$_{4.23}$ & 3.48$_{4.26}$ & 3.48$_{3.49}$ & 3.49$_{3.07}$ \\
    \midrule
    \multicolumn{11}{c}{\textit{Wisconsin (WI)}} \\
    \midrule
    PD & 3.98$_{4.12}$ & 4.47$_{4.37}$ & 3.68$_{4.24}$ & 2.91$_{2.52}$ & 3.84$_{4.10}$ & 3.77$_{4.46}$ & 2.63$_{3.56}$ & 3.09$_{2.44}$ & 3.17$_{3.30}$ & 3.40$_{3.75}$ \\
    PPS & 1.31$_{1.52}$ & 9.57$_{1.32}$ & 9.55$_{9.19}$ & 9.58$_{1.05}$ & 9.55$_{1.24}$ & 1.21$_{1.17}$ & 9.43$_{9.18}$ & 9.52$_{1.04}$ & 9.60$_{9.70}$ & 9.55$_{1.00}$ \\
    Bias & 3.56$_{3.57}$ & 2.73$_{3.29}$ & 1.26$_{1.21}$ & 2.29$_{2.27}$ & 1.67$_{2.08}$ & 3.65$_{4.54}$ & 2.33$_{2.16}$ & 1.45$_{1.84}$ & 1.45$_{1.97}$ & 1.94$_{1.48}$ \\
    Unfairness & 4.95$_{5.56}$ & 3.94$_{4.10}$ & 3.93$_{3.01}$ & 3.94$_{3.22}$ & 3.93$_{4.73}$ & 4.94$_{6.09}$ & 3.91$_{3.76}$ & 3.93$_{3.84}$ & 3.92$_{4.16}$ & 3.92$_{4.28}$ \\
    \bottomrule
    \end{tabular}
    \end{table*}

\subsection{Heterogeneous-Agent Configurations and Model Asymmetry}

In real-world scenarios, it is unlikely that both political parties involved in negotiations would operate under equal conditions or assumptions, particularly with regard to the capabilities of the models they use. To address this potential source of bias, we conducted an experiment exploring the effects of using heterogeneous-agent configurations. This study involved the use of models with varying capabilities for the Democratic and Republican agents across seven key swing states: Arizona (AZ), Georgia (GA), Michigan (MI), North Carolina (NC), Nevada (NV), Pennsylvania (PA), and Wisconsin (WI). In this experiment, the agents were assigned different models representing various performance tiers. 

\begin{table*}[htp]
\centering
\begin{tabular}{c|ccccccc}
\toprule
\textbf{Setting} & \textbf{Dem} & \textbf{Rep} & \textbf{GA} & \textbf{PA} & \textbf{MI} & \textbf{AZ} & \textbf{NV} \\
\midrule
1 & GPT-5 & Gemini-2.5-pro & 0.3482 & 0.3494 & 0.3994 & 0.4086 & 0.4390 \\
2 & GPT-5 & Gemini-2.0-flash-lite & 0.3422 & 0.3482 & 0.3866 & 0.3992 & 0.4338 \\
3 & GPT-4o-mini & Gemini-2.0-flash-lite & 0.3492 & 0.3494 & 0.4011 & 0.4013 & 0.4561 \\
4 & Gemini-2.5-pro & GPT-5 & 0.3459 & 0.3489 & 0.4005 & 0.4097 & 0.4450 \\
5 & Gemini-2.5-pro & GPT-4o-mini & 0.3523 & 0.3554 & 0.4063 & 0.4025 & 0.4457 \\
6 & Gemini-2.0-flash-lite & GPT-4o-mini & 0.3389 & 0.3352 & 0.3853 & 0.3954 & 0.4361 \\
CD\_2020 &  &  & 0.3559 & 0.3559 & 0.4151 & 0.4240 & 0.4564 \\
\bottomrule
\end{tabular}
\caption{Unfairness metrics for various agent configurations across seven states.}
\label{tab:asymmetric_results}
\end{table*}
\paragraph{Results.}
The results, summarized in Table \ref{tab:asymmetric_results}, show that even under conditions where the two parties used agents with differing capabilities, the fairness of the negotiation process remained stable. Across all configurations, the unfairness metric was consistently lower than the baseline of CD\_2020. However, in configurations where there was a clear disparity in model capabilities (e.g., when one party used a significantly more capable model, as in Setting 5), there was a slight increase in unfairness, which aligns with expectations.

These findings suggest that the Choose-and-Freeze protocol maintains its self-stabilizing nature even when the agents involved are not equally matched in terms of their computational power. The protocol's fairness is therefore primarily driven by the structural dynamics of the negotiation process rather than by an idealized assumption of equal agent capabilities.

\subsection{Case Study}
\label{app:case}
\begin{figure*}[ht]
\centering
\includegraphics[width=\textwidth]{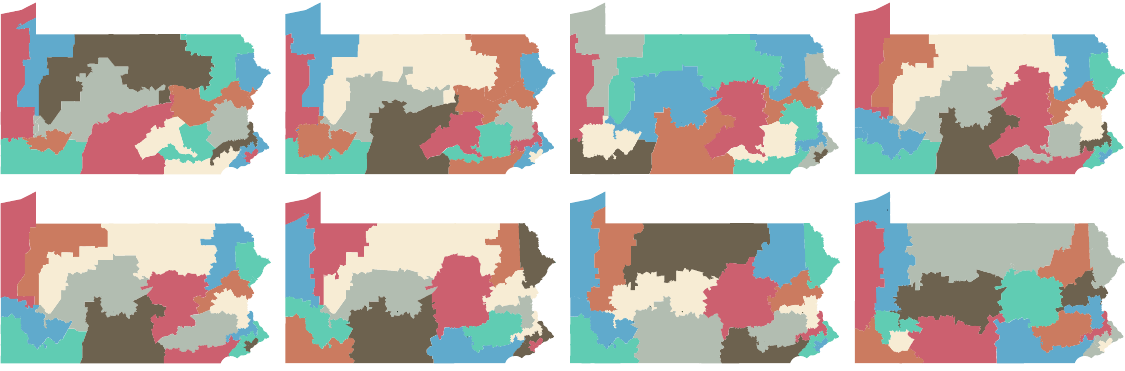}
\caption{Eight districting plans generated by Agentmandering in Pennsylvania.}
\label{fig:map}
\end{figure*}

In Pennsylvania, we visualize the districting decisions made by Agentmandering. Figure~\ref{fig:map} shows eight districting plans generated across different runs of the algorithm. The results appear relatively stable, with districts preserving geographic coherence and avoiding extreme distortions. While some local variations exist, the overall shapes and partisan compositions remain consistent, demonstrating that the \textit{Choose-and-Freeze} mechanism—when guided by reasoning-capable agents—yields reproducible and interpretable outcomes.

Notably, several districts remain nearly identical across all plans, indicating a high degree of structural convergence. In particular, the Philadelphia metropolitan area consistently forms a compact cluster of urban-majority districts, while the western and central rural regions often coalesce into geographically contiguous Republican-leaning zones. Analysis of the agent decisions reveals that these districts are frequently selected or frozen early in the process, suggesting that both agents quickly recognize their strategic stability. This behavior highlights the effectiveness of the \textit{Choose-and-Freeze} protocol in reinforcing durable geographic and demographic boundaries under competitive interaction.

\bibliography{references}

\end{document}